\documentclass[]{ceurart}

\usepackage{listings}
\usepackage{multirow}
\usepackage{longtable}
\usepackage{enumitem}

\begin{document}

\copyrightyear{2026}
\copyrightclause{Copyright for this paper by its authors.
  Use permitted under Creative Commons License Attribution 4.0
  International (CC BY 4.0).}

\conference{CLEF 2026 Working Notes, 21-24 September 2026, Jena, Germany}
           
\title{DS@GT ARC at eRisk 2026: Hybrid Multi-Agent LLM System with Structured Algorithmic Guidance for Conversational Depression Screening}

\title[mode=sub]{Notebook for eRisk Lab's Task 1 in CLEF 2026}

\author[1]{Victor Gong}[%
orcid=0009-0002-1080-5980,
email=vgong7@gatech.edu
]
\cormark[1]

\author[1]{David Guecha}[orcid=0009-0009-9855-5330,
email=dahumada3@gatech.edu
]

\address[1]{Georgia Institute of Technology, North Ave NW, Atlanta, GA 30332}
\cortext[1]{Corresponding author.}

\begin{abstract}
  We describe DS@GT's submission to the eRisk 2026 Task 1 challenge on
  conversational depression screening, in which systems interview LLM
  personas that simulate individuals with varying depression profiles and
  produce a Beck Depression Inventory II (BDI-II) score plus four key symptoms per persona, without directly asking sensitive mental health questions. Our pipeline evolved through three stages: a monolithic single-model prototype to start off, a baseline multi-agent architecture that separates conversational interviewing from BDI-II scoring under a coordinating orchestration layer, and a final hybrid configuration that replaces the paid GPT-5-nano interviewer with the open-source Gemma 27B. To offset the model's weaker reasoning and instruction-following, the hybrid adds three algorithmic components: a precomputed dialogue tree that standardizes interview openers and follow-ups, a reliability-weighted consensus aggregation inspired by the Weaver framework, and a cluster-based imputation step for unprobed symptoms. We submitted three fully automated runs across all 20 personas, with Run 1 from the paid baseline and Runs 2 and 3 from the hybrid. Hybrid Run 3 achieved an ADODL of 0.9063, ranking 3rd among all complete-submission runs and placing DS@GT 2nd among the 21 teams overall, while outperforming our paid baseline Run 1 (0.8841) at roughly one-quarter of the per-persona API cost. These results support our central hypothesis that with sufficient algorithmic supervision, a weaker open-source model can compete with a stronger proprietary model in the conversational interviewer role. Our source code is available at https://github.com/dsgt-arc/erisk-task1-2026.
\end{abstract}


\maketitle

\section{Introduction}

Depression affects over 280 million people worldwide~\cite{cite_depression} and frequently goes undetected until symptoms reach clinical severity~\cite{faisalcury2022}. Structured instruments such as the Beck Depression Inventory II (BDI-II)~\cite{beck1996} provide a validated framework for severity assessment, but their administration requires trained clinicians, limiting scalability in resource-constrained settings. Automated conversational systems that can elicit and interpret depressive symptom information represent a practical direction for broadening screening access, provided they can produce reliable assessments without direct clinical inquiry. Recent work supports this direction: systematic reviews report strong LLM performance across depression-related tasks~\cite{omar2025}, depression symptoms including suicidal ideation are observable in online social media discourse~\cite{aragon2025}, and structuring clinical interview content as directed graphs for LLM-based evaluation has produced competitive automated depression detection~\cite{chen2024depression}.

The eRisk 2026 Task 1 challenge provides a controlled testbed for this problem. Participants interact conversationally with LLM personas, LoRA adapters fine-tuned on Llama-3-8B-Instruct to simulate individuals with varying depression profiles, and must estimate a BDI-II score and
identify up to four prominent symptoms per persona. The task prohibits direct questions about depression and requires strategies to infer symptom severity from natural conversation. Evaluation is also penalized for late decisions through the number of conversation turns, which creates a trade-off between thorough diagnosis and conversational efficiency.

Two characteristics of the task make standard modeling and development difficult. First, ground truth labels are not provided during the entirety of the submission period, meaning systems must operate ``in the dark'' and predictions can't be directly tuned against task-specific labels.
Second, personas are released weekly with hard submission deadlines, requiring a system that is both reliable from the outset and improvable across the whole period. Under these constraints, we adopted an iterative development strategy where each week's persona helped validate and test design changes.

Building on our prior DS@GT eRisk 2025 submission~\cite{miyaguchi2025}, our system evolved through three stages. An initial single-model prototype established end-to-end functionality but exhibited score drift and inconsistent symptom coverage across repeated runs. A multi-agent architecture separating conversational interviewing from BDI-II
scoring improved coverage and reliability, at the cost of roughly double the API expenditure. A final hybrid configuration retained the multi-agent structure but replaced the paid interviewer model with the open-source Gemma 27B, compensating for the model's looser adherence and higher variance through algorithmic supervision: a precomputed dialogue tree, reliability-weighted sample aggregation, and imputation across symptom clusters. For the majority of the personas, we employed the paid configuration for Run 1 and hybrid configuration for Runs 2 and 3.

The central hypothesis motivating this work is that a weaker, open-source model can produce results competitive with a stronger paid model in the interviewer role, given sufficient algorithmic guidance. Our results support this claim. In the official evaluation, both hybrid-produced Run 2 and Run 3 outperformed our paid-model Run 1 in ADODL, the primary evaluation metric, at approximately one-quarter of the per-persona API cost. Furthermore, Run 3 placed 3rd overall among all complete-submission runs and 2nd among all teams.

\section{Task and Evaluation}

Participants interact with up to 20 LLM personas released weekly. Each persona is implemented as a LoRA adapter on Llama-3-8B-Instruct and simulates an individual with a specific depression profile. For each persona, participants submit a BDI-II score and up to four key symptoms. Up to three fully automated runs are permitted per persona.

The BDI-II questionnaire covers 21 symptoms, such as sadness, pessimism, self-criticalness, sleep changes, fatigue, and loss of concentration, each scored 0-3 for a maximum total of 63. Severity is conventionally categorized as minimal (0-13), mild (14-19), moderate (20-28), and severe (29-63).

\section{System Architecture \& Methods}

\subsection{Monolithic Single-Model Prototype}

Our initial prototype assigned all conversational and scoring responsibilities to a single language model with a unified context-heavy prompt. The system conducted a multi-turn interview with the
LLM persona, then produced a structured BDI-II score report at the conclusion of the conversation. Prompt engineering focused on two priorities: guiding the model to target related symptom clusters within single turns to reduce total turn count, and providing explicit 0-3
scoring rubrics for each BDI-II symptom to improve inter-run consistency.

While this approach confirmed end-to-end functionality, it exhibited two systematic failure modes. First, assigning both interviewing and scoring to a single model produced score drift during preliminary experimentation. Repeated runs on identical personas yielded substantially different BDI-II totals and symptom rankings, reflecting sensitivity to sampling stochasticity in both the interviewer and the persona. Second, without an external mechanism enforcing symptom coverage, the model would sometimes exhaust its turn budget on a subset of symptoms, leaving others unprobed. These instabilities were compounded by non-determinism in the persona itself. Persona 3, for example, responded to the same probe (``Have you been harder on yourself than usual?'') with ``I don't think I'm actually hard on myself'' in one session and ``definitely harder on myself'' in another, introducing variance in conversation and scoring that is hard to control with a single-model system.

\subsection{Multi-Agent Architecture (Baseline)}

\begin{figure}[!htbp]
  \centering
  \includegraphics[width=\linewidth]{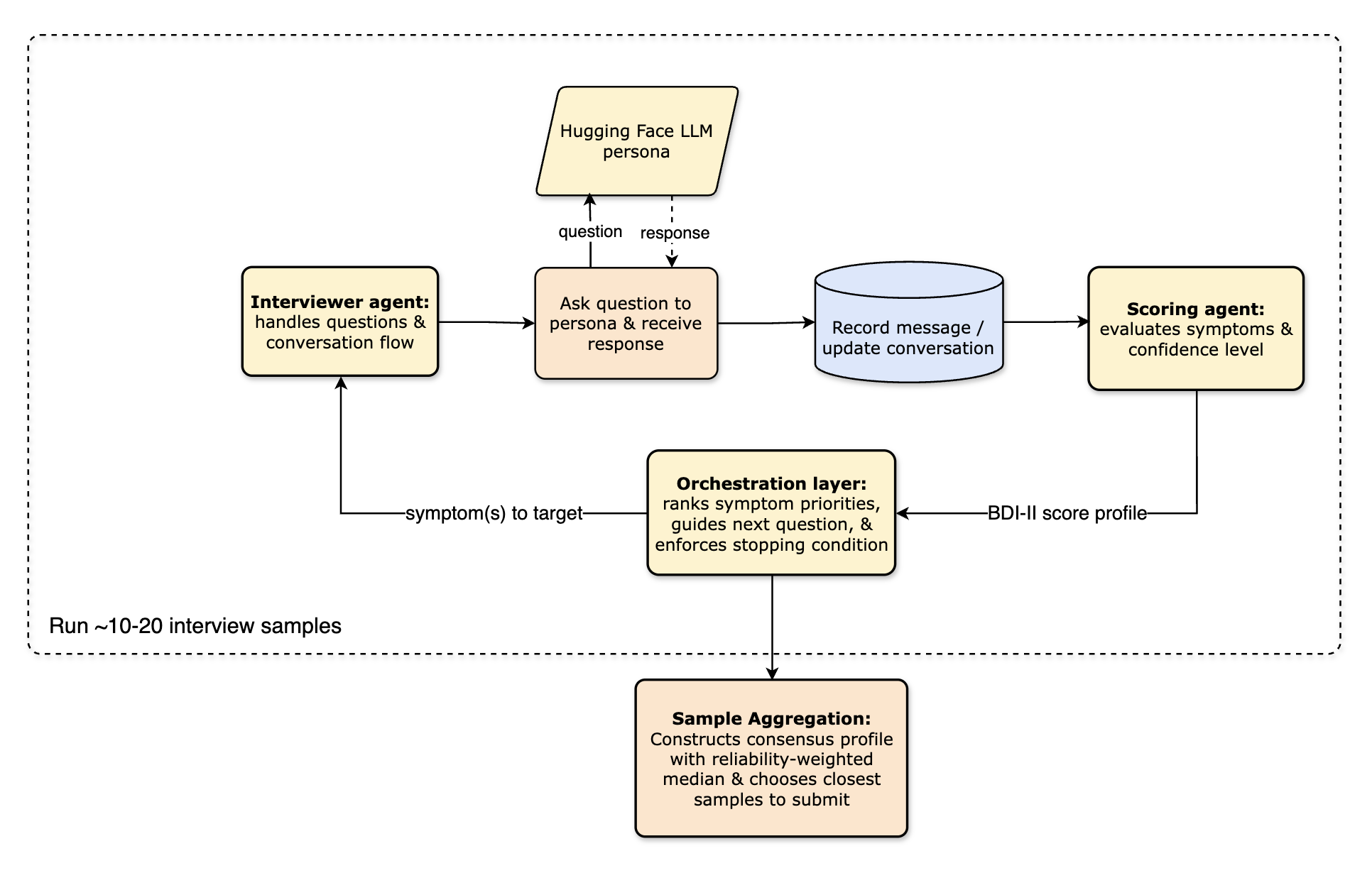}
  \caption{Multi-agent system architecture. The interviewer agent conducts turn-by-turn dialogue with the LLM persona, and the scorer agent updates BDI-II symptom estimates and confidence levels after each turn. The orchestration layer ranks symptom priorities, directs the interviewer toward the highest-priority targets, and enforces the stopping conditions. The full loop runs 10 to 20 interview samples per persona, and the sample aggregation step constructs a consensus profile via reliability-weighted median to select the closest runs for submission.}
  \label{fig:architecture}
\end{figure}

To address the reliability limitations of the monolithic system, we separated conversational and scoring responsibilities into two specialized agents coordinated by an orchestration layer (Figure~\ref{fig:architecture}). This approach was inspired by team SINAI's similar multi-agent design from eRisk 2025~\cite{cite_erisk2025sinai}.

The \textbf{interviewer agent} conducts adaptive multi-turn conversations with the persona. Its prompt instructs it to ask targeted, naturalistic questions about potential depressive symptoms, pursue severity-distinguishing follow-ups when initial responses are ambiguous, and avoid direct questions about depression or mental health.

The \textbf{scorer agent} operates in parallel with the interview, reviewing the full conversation transcript after each turn and producing updated 0-3 estimates for all 21 BDI-II symptoms along with per-symptom confidence scores. By running the scorer at every turn rather than at the
conversation end, the orchestration layer receives a continuously updated diagnostic picture.

The \textbf{orchestration layer}, consistent with recent findings that a centralized orchestrator improves coordination in multi-agent LLM systems~\cite{dang2025orchestration}, serves three functions. First, it groups the 21 BDI-II symptoms into seven thematic clusters and ranks them based on a priority score combining current estimated severity and inverse confidence to direct the interviewer toward the least diagnosed symptoms at each turn. Second, it embeds cluster-specific instructions into the interviewer's prompt dynamically, so the conversational agent addresses high-priority areas. Third, it enforces stopping conditions based on four criteria evaluated after each turn: (a) the 18-turn hard limit, (b) all symptoms exceeding 0.6 confidence after more than 8 turns, (c) the BDI-II total changing by fewer than 2 points over the last 3 consecutive turns after more than 10 turns, or (d) at least 70\% of symptoms probed, subject to the 18-turn override. These conditions help minimize the number of turns per conversation and enforce the hard limit imposed by the task. We selected the numerical thresholds (the turn caps, the 0.6 confidence floor, and the 70\% coverage criterion) heuristically, tuned by manual inspection on the released personas in the absence of ground truth during the submission period.

To mitigate run-level variance, we batched 30 interview samples per persona and selected the three runs whose total BDI-II score fell closest to the batch median for submission. We later kept this initial version of our multi-agent system as our \textit{baseline configuration} for the remainder of the project.

\subsection{Hybrid Configuration}

The multi-agent architecture improved coverage and consistency but approximately doubled API cost relative to the monolithic prototype, due to the passage of full conversational histories to two model instances per turn. This motivated an attempt to achieve similar predictive results
with a cheaper, open-source model. We retained GPT-5-nano~\cite{gpt5nano} for the scorer, where instruction-following precision directly affects diagnostic accuracy, while replacing it with Gemma 27B~\cite{gemma3} for the interviewer, where conversational naturalness and question diversity are the primary requirements. All other components of the system (the scorer agent, the orchestration layer, the stopping conditions, and the multi-agent loop itself) were held constant or improved in parallel between the baseline and hybrid.

Initial results with this hybrid configuration revealed two predictable failure modes consistent with weaker open-source models. The hybrid systematically underestimated BDI-II totals, particularly for personas with prominent cognitive and affective symptoms, and exhibited higher score variance across runs than the paid baseline. We attribute both effects to Gemma 27B's tendency to accept surface-level responses without probing for underlying emotional content, and to less reliable adherence to the structured interview instructions embedded in the prompt~\cite{cite_hallucination}. We developed three algorithmic components to address these limitations.

\begin{figure}[!htbp]
  \centering
  \includegraphics[width=0.85\linewidth]{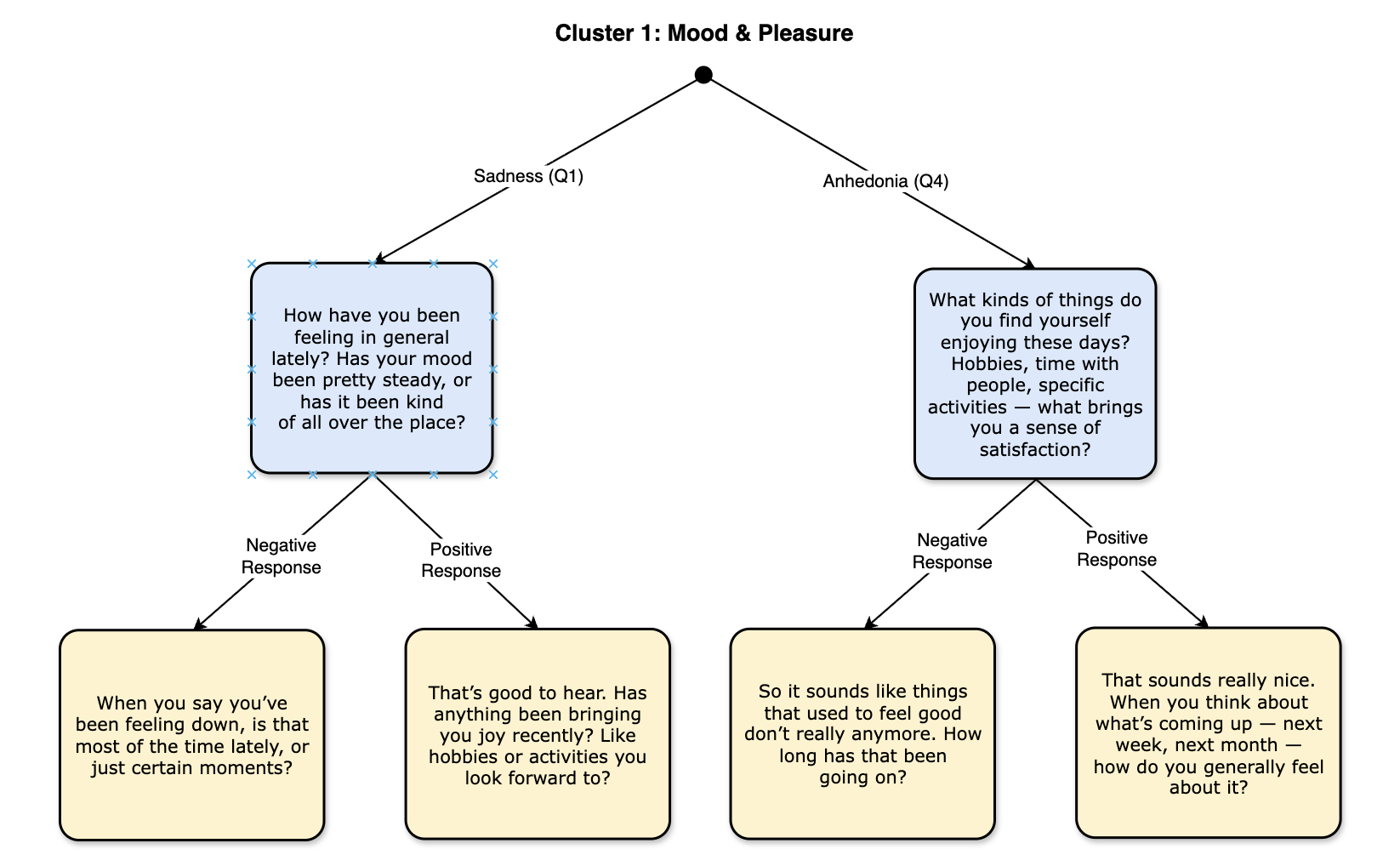}
  \caption{Cluster 1 (Mood \& Pleasure) of the precomputed dialogue tree, illustrating the two opener questions (sadness Q1 and anhedonia Q4) and their negative- and positive-response follow-up branches. The complete tree spanning all seven clusters appears in Appendix~\ref{appendix:dialogue_tree}.}
  \label{fig:dialogue_tree}
\end{figure}

\paragraph{Dialogue Tree.} Structured task frameworks have been shown to improve controllability and coherence in LLM dialogue agents~\cite{li2024chatsop}. To standardize interview openers and reduce the variance introduced by unconstrained question generation, we constructed a precomputed dialogue tree with 7 symptom clusters, 12 cluster-opening questions, and 17 follow-up probes. Symptoms are grouped thematically so that semantically related items can be probed within a single conversational thread, with each individual symptom or small symptom pair assigned its own dedicated opener. Grouping symptoms this way exploits natural conversational flow, since related symptoms tend to co-occur in the same emotional register. A persona already discussing energy and fatigue is more likely to respond openly to a follow-up about sleep than to an abrupt topic shift, producing richer responses and more reliable scorer estimates. Clinically sensitive symptoms, such as suicidal ideation and loss of interest in sex, have no precomputed lines so as to treat them on a case-by-case basis.

We hardcoded the very first question of the conversation to target sadness and anhedonia. Subsequent openers are selected on demand, where the system retrieves the first unused opener whose target symptoms appear in the orchestrator's current priority list of unassessed or low-confidence items. As the scorer gains confidence across turns, symptoms drop out of the priority list and their openers are skipped, so not every opener fires in every interview. Within each opener, BM25~\cite{robertson2009} text matching on the persona's response chooses the best follow-up branch. This structure covers approximately 2-3 interview turns without any LLM call (Figure~\ref{fig:dialogue_tree}).

\paragraph{Reliability-Weighted Aggregation.} Simple median selection, as used in the multi-agent baseline, is sensitive to outlier runs when per-sample variance is high. Drawing on the classical ensemble learning principle that many weak predictors combined intelligently can match a single
strong one, recently adapted to LLM verifier aggregation by the Weaver framework~\cite{cite_weaver}, we replaced it with a reliability-weighted consensus method. For each pair of runs $(a,b)$ and each BDI-II symptom $s$, we compute a pairwise agreement score:

\begin{equation}
  A(a, b, s) = 1 - \frac{|score_{s}^{a} - score_{s}^{b}|}{3}
\end{equation}

The reliability weight for run $r$ is the mean pairwise agreement across all other runs in the batch and all 21 symptoms. A consensus BDI-II profile is constructed as the reliability-weighted median per symptom, and the three runs with minimum L1 distance between their total score and the consensus total are selected for submission. This method identifies internally consistent samples without requiring ground truth, effectively using the batch itself as a weak supervision signal.

\paragraph{Cluster Imputation.} Because Gemma 27B's shallower probing sometimes left symptoms entirely unaddressed within the 18-turn limit, unprobed symptoms would contribute a default score of zero, systematically depressing BDI-II totals. To address this, we assigned each of the 21 BDI-II symptoms three semantically related donor symptoms, manually chosen by similarity, from which its severity can be inferred. If all three donors have estimated severity greater than 2 and confidence greater than 0.7, the unprobed symptom is assigned the mean of the donor scores, capped at 2. This prevents systematic undershooting while remaining conservative. Imputed scores are never assigned at maximum severity, and imputation is only triggered when donor evidence is strong and consistent. Like the orchestration thresholds, we chose the 0.7 confidence floor as a reasonable estimate in light of no ground truth.

\medskip

For the hybrid configuration, we ran 20 interviews per persona versus 10 for the baseline. This asymmetry directly compensates for higher per-sample variance in the weaker interviewer and is grounded in the weak supervision principle that motivates Weaver's design, where combining a larger pool of weak verifiers can match the reliability of fewer strong ones. Because one hybrid interview costs approximately half the price of a paid-model interview, running 20 hybrid samples is cost-equivalent to the 10-run baseline while providing the reliability aggregation with substantially denser pairwise signal (190 comparisons versus 45). The larger pool is not a
methodological concession but the intended operating mode of the approach.

\section{Results \& Discussion}

\subsection{Baseline vs.\ Hybrid Performance Comparison}

Because ground truth BDI-II scores were not released during the submission period, we evaluated the hybrid configuration against the paid-model baseline through internal comparison across Personas 7 to 20, the 14 personas for which both systems were run in parallel.

The most direct advantage of the hybrid is cost. Substituting Gemma 27B for the GPT-5-nano interviewer reduced per-persona API cost from approximately \$8 to \$2, a 75\% reduction.

\begin{figure}[htbp]
  \centering
  \includegraphics[width=0.7\linewidth]{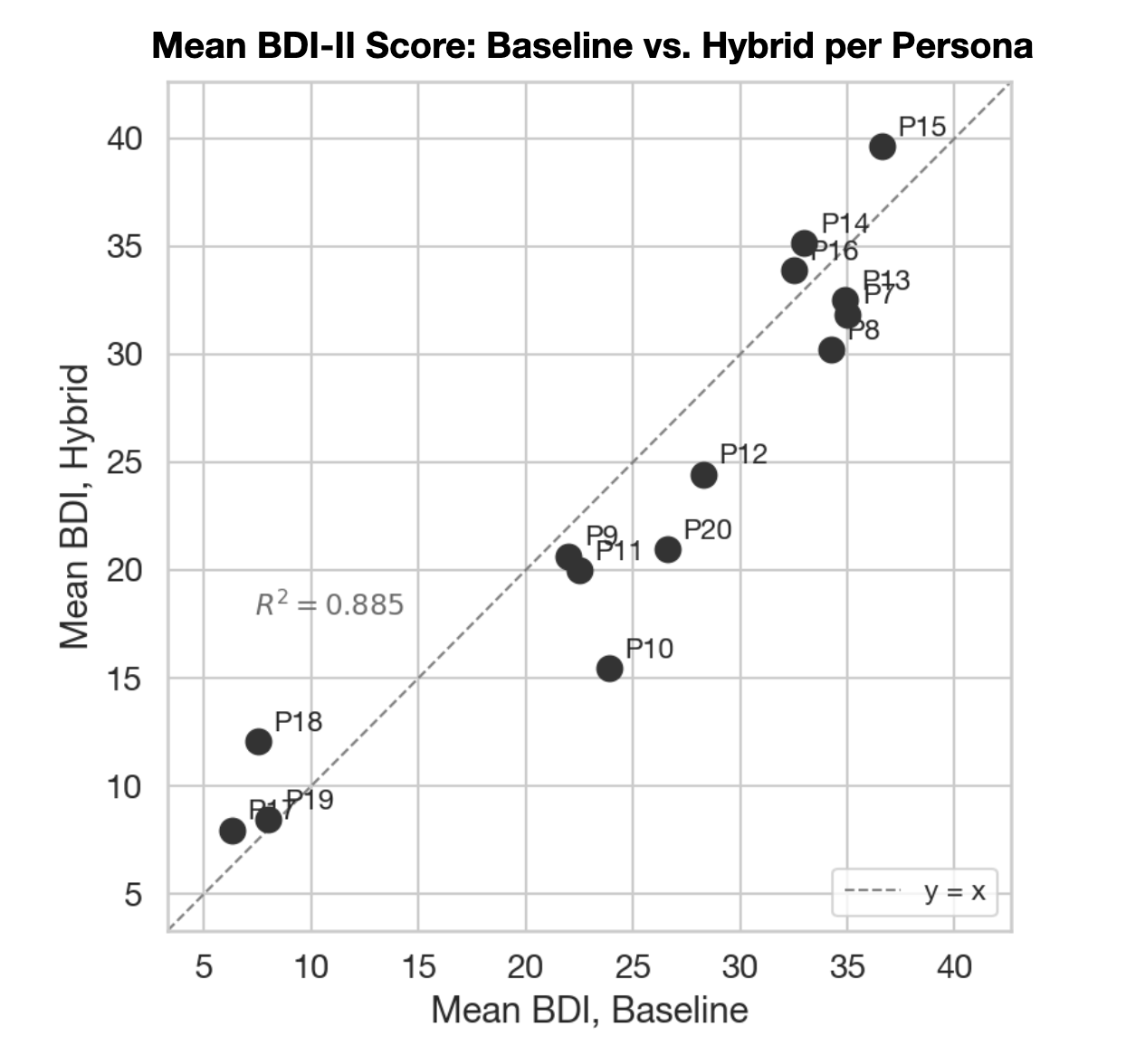}
  \caption{Per-persona mean BDI-II score comparison between the baseline (GPT-5-nano interviewer, 10 runs) and hybrid (Gemma 27B interviewer, 20 runs) configurations across Personas 7 to 20. The dashed line indicates $y = x$ (perfect agreement), and the Pearson coefficient of determination is $R^2 = 0.885$. Points below the diagonal correspond to personas where the hybrid produced lower mean BDI-II estimates than the baseline, with Persona 10 the most pronounced outlier.}
  \label{fig:scatter}
\end{figure}

The hybrid underestimated baseline BDI-II totals for 8 of the 14 personas assessed (Figure~\ref{fig:scatter}). This directional bias is consistent with Gemma 27B's tendency toward surface acceptance rather than affective probing~\cite{cite_hallucination}. Despite this offset, most of the hybrid's
scores still cluster near the diagonal line in Figure~\ref{fig:scatter}, which indicates competitive score alignment across the board.

Quantitatively, the hybrid's mean scores remained within 6 BDI-II points, approximately one severity category, of the baseline for 13 of 14 personas (Table~\ref{tab:personas}). The single exception was
Persona 10, where the hybrid underestimated by a mean of 8.4 points (23.9 baseline vs.\ 15.4 hybrid). Analysis of the submitted runs for Persona 10 reveals a qualitatively different symptom profile. The baseline run (Run 1), scoring 23, identified cognitive and self-evaluative symptoms (pessimism, guilty feelings, self-dislike, self-criticalness), while both hybrid runs (Runs 2 and 3), scoring 20 and 16 respectively, anchored on somatic symptoms (sleep changes, fatigue, loss of energy, concentration difficulty) without probing the underlying affective layer, producing a lower total score.

\begin{table}[htbp]
  \caption{Mean BDI-II scores for the baseline (GPT-5-nano interviewer, 10 runs) and hybrid (Gemma 27B interviewer, 20 runs) systems across Personas 7-20. Gap denotes absolute difference in means. Persona 10 is the only case exceeding one BDI-II severity category (gap $>$ 7).}
  \label{tab:personas}
  \begin{tabular}{lccc}
    \toprule
    Persona & Baseline Mean & Hybrid Mean & Gap \\
    \midrule
    7           & 35.0 & 31.8 & 3.1 \\
    8           & 34.2 & 30.2 & 4.1 \\
    9           & 22.0 & 20.6 & 1.4 \\
    \textbf{10} & \textbf{23.9} & \textbf{15.4} & \textbf{8.4} \\
    11          & 22.5 & 20.0 & 2.5 \\
    12          & 28.3 & 24.4 & 3.9 \\
    13          & 34.9 & 32.5 & 2.4 \\
    14          & 33.0 & 35.2 & 2.1 \\
    15          & 36.6 & 39.6 & 3.0 \\
    16          & 32.5 & 33.9 & 1.4 \\
    17          & 6.3  & 8.0  & 1.7 \\
    18          & 7.5  & 12.0 & 4.6 \\
    19          & 8.0  & 8.4  & 0.4 \\
    20          & 26.6 & 21.0 & 5.7 \\
    \bottomrule
  \end{tabular}
\end{table}

\begin{figure}[!htbp]
  \centering
  \includegraphics[width=\linewidth]{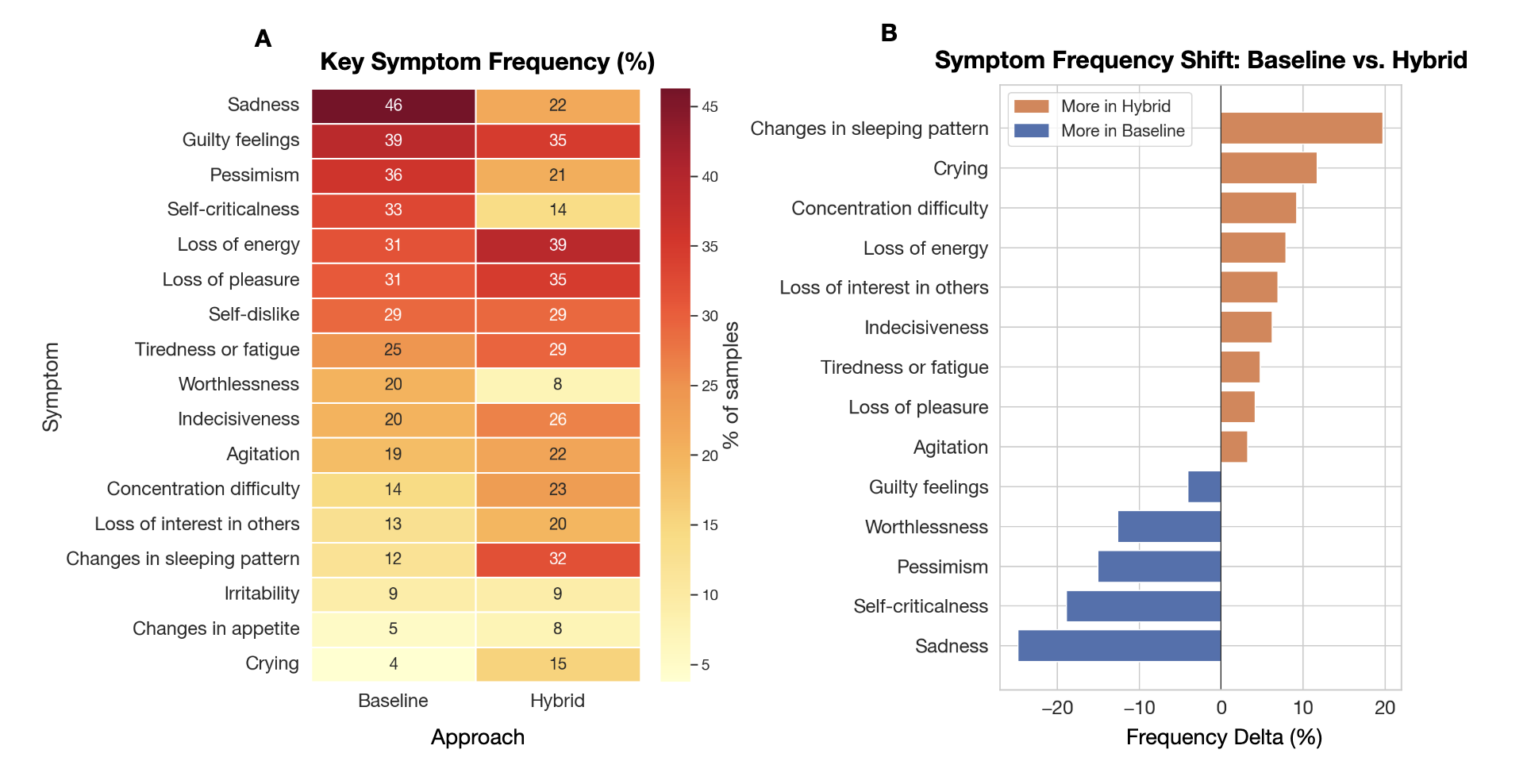}
  \caption{Symptom flagging frequency across Personas 7 to 20 for the baseline and hybrid configurations. (A) Heatmap of \% samples in each pool that listed the symptom as one of the four key symptoms. (B) Difference in symptom hit rates between hybrid and baseline. The baseline model tended to focus on intrinsic, emotional symptoms like sadness and guilt, while the hybrid favored surface-level somatic signs like sleep and fatigue.}
  \label{fig:symptoms}
\end{figure}

Examining tendencies at the symptom level further reinforces this trend. The hybrid overrepresented somatic symptoms like fatigue, sleep disturbance, and loss of energy, while underrepresenting affective symptoms such as sadness and anhedonia. This asymmetry reflects a structural tendency in Gemma 27B to treat somatic framing at face value rather than pursue the emotional content that underlies it, a limitation the dialogue tree's affective probing
branches only partially addressed (Figure~\ref{fig:symptoms}).

\begin{figure}[!htbp]
  \centering
  \includegraphics[width=\linewidth]{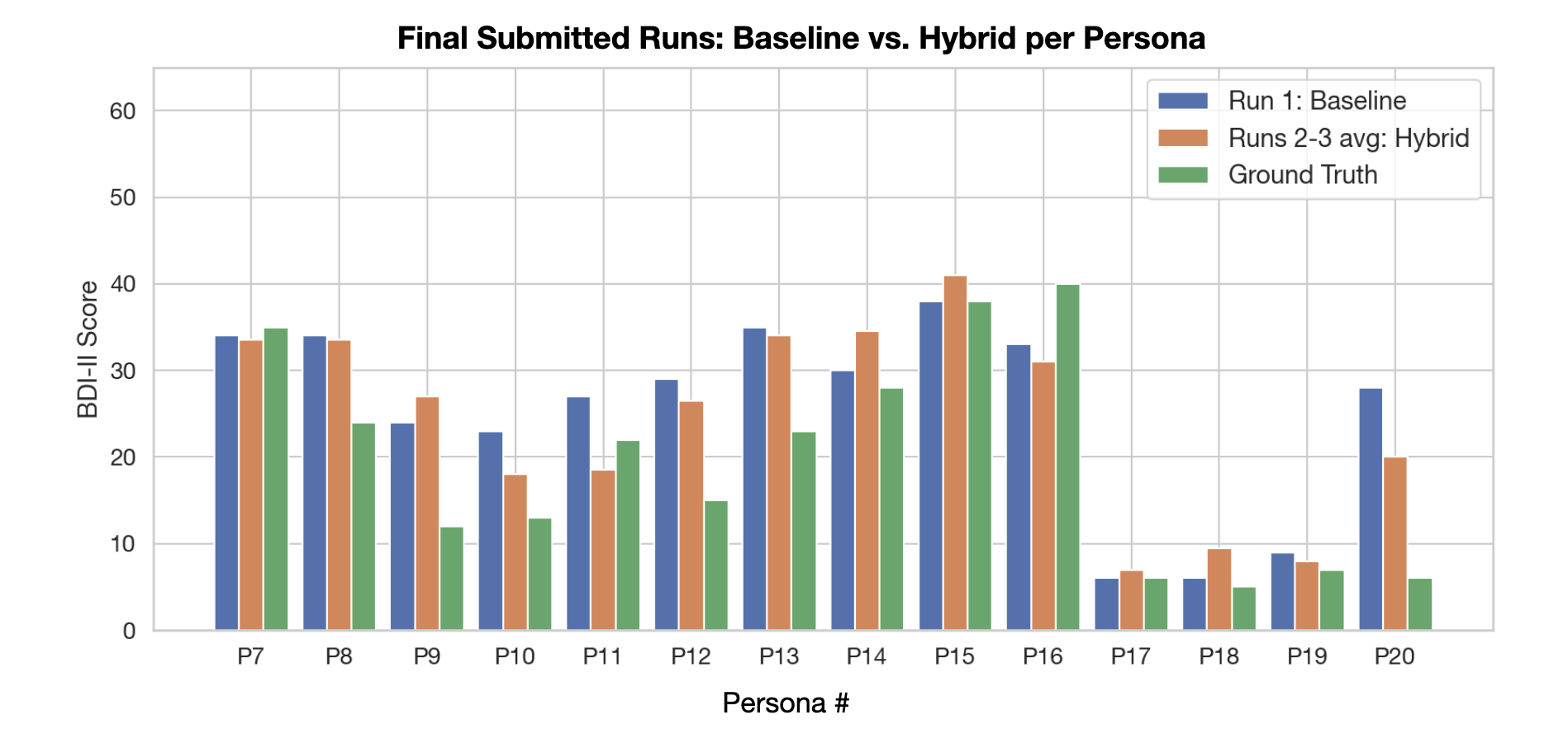}
  \caption{Final submitted runs per persona compared against ground truth. Blue bars show the BDI-II score of the baseline-derived Run 1 (GPT-5-nano interviewer). Orange bars show the mean of the two hybrid-derived submissions, Runs 2 and 3 (Gemma 27B interviewer). Green bars show each persona's ground-truth BDI-II score. Most personas show close agreement between the two configurations, with the largest gaps at Personas 10, 11, and 20.}
  \label{fig:runs}
\end{figure}

Nevertheless, runs from both pools converged to comparable scores across nearly all personas (Figure~\ref{fig:runs}). Personas 11 and 20 showed larger discrepancies than the others, differences that were less apparent in the mean scores. These gaps likely stem from differences in aggregation methods, simple median in the baseline versus reliability-weighted aggregation in the hybrid, rather than weaker model performance. Measured against the ground-truth BDI-II scores (green bars), both configurations land near the true value for many personas but share a tendency to overestimate severity on a subset of cases. The clearest example is Persona 20, whose true minimal-range score of 6 was scored in the moderate range by both the baseline (28) and the hybrid (20), with similar joint overestimates for Personas 8, 9, 12, and 13. Because these errors move together across the paid and free configurations rather than separating them, we attribute them to the difficulty of eliciting candid disclosure from deflecting personas rather than to the open-source substitution. Overall, the small margins between the two systems support our hypothesis that with sufficient supervision and effective aggregation, the weaker interviewer model in the hybrid can perform competitively with the paid baseline.

\subsection{Official Judging Results}

The official eRisk 2026 evaluation results corroborate these internal findings. DS@GT submitted three fully automated runs across all 20 personas. Run 1 was drawn from the baseline configuration (GPT-5-nano interviewer), and Runs 2 and 3 from the hybrid (Gemma 27B interviewer). Our system averaged 11.5 messages per conversation at 210.15 characters per
message, slightly higher and denser than most other teams~\cite{PerezEtAl2026eRiskWorkingNotes}~\cite{PerezEtAl2026eRiskLNCS} (Table~\ref{tab:teams}).

\begin{table}[!htbp]
  \caption{Conversation statistics for selected eRisk 2026 Task 1 teams. \emph{msgs./conv.} denotes the average number of system messages per persona interview. \emph{chars./msg.} denotes the average characters per system message.}
  \label{tab:teams}
  \begin{tabular}{lrr}
    \toprule
    Team & msgs./conv. & chars./msg. \\
    \midrule
    DS@GT           & 11.5 & 210.15 \\
    pjmathematician & 6.0  & 108.88 \\
    SINAI           & 4.4  & 242.61 \\
    Awakened        & 5.3  & 84.37  \\
    UNED-GELP       & 10.7 & 114.47 \\
    \bottomrule
  \end{tabular}
\end{table}

\medskip

Submissions were judged on several metrics that encapsulate the accuracy and efficiency of their BDI-II score and symptom predictions:

\begin{itemize}
    \item \textbf{Depression Category Hit Rate (DCHR)}: Fraction of runs whose BDI-II scores fell into the correct depression category (score range).
    \item \textbf{Average Difference between Overall Depression Levels (ADODL)}: Average distance between estimated depression level (BDI-II score) and actual depression level, expressed as $(63-|ADL-EDL|)/63$ and averaged across all runs. This benchmark served as the primary evaluation metric.
    \item \textbf{Average Symptom Hit Rate (ASHR)}: Ratio of correctly identified key symptoms averaged across all runs. Due to the ambiguity and closeness of some symptoms, pushing this quantity past 0.5 (2/4 symptoms correct) proved to be difficult across all teams.
    \item \textbf{Latency-Aware DCHR and ASHR (LDCHR, LASHR)}: DCHR and ASHR with an additional ``decay'' coefficient that penalizes long conversations, taking into account the number of turns used.
\end{itemize}

Across the 21 complete-submission teams, DS@GT ranked 2nd overall by ADODL, with Run 3 achieving a score of 0.9063, behind only pjmathematician's best run (0.9254). Notably, both hybrid runs outperformed the baseline Run 1 on ADODL, which directly supports our central hypothesis that with sufficient algorithmic supervision, the open-source Gemma 27B interviewer can match and even exceed the paid GPT-5-nano baseline on the primary evaluation metric (Table~\ref{tab:adodl}).

However, this improvement didn't extend to all metrics. Run 3's ASHR of 0.1875 fell below the baseline's 0.2500, while Run 2 matched the baseline at 0.2500. This degradation aligns with the earlier observation that the hybrid configuration overemphasizes certain somatic symptoms at the expense of deeper affective symptoms (Figure~\ref{fig:symptoms}), leading to improved BDI-II scoring but weaker symptom identification.

\begin{table}[!htbp]
  \caption{Best-performing submitted run per team across the five eRisk 2026 evaluation metrics: DCHR (Depression Category Hit Rate), ADODL (Average Difference Between Overall Depression Levels), ASHR (Average Symptom Hit Rate), LDCHR (Latency-adjusted DCHR), and LASHR (Latency-Adjusted Symptom Hit Rate). Parenthesized numbers indicate each team's rank on that metric across all 21 complete-submission teams. Bold ADODL values highlight DS@GT Run 3's third-place finish.}
  \label{tab:adodl}
  \small
  \begin{tabular}{lccccc}
    \toprule
    Team & DCHR & ADODL & ASHR & LDCHR & LASHR \\
    \midrule
    pjmathematician Run 3 & 0.5500 (2) & \textbf{0.9254 (1)} & 0.3000 (4)  & \textbf{0.3872 (1)} & 0.2112 (3)  \\
    DS@GT Run 3           & 0.5500 (2) & \textbf{0.9063 (3)} & 0.1875 (11) & 0.2493 (6)          & 0.0777 (14) \\
    SINAI Run 3           & 0.4737 (4) & 0.8989 (7)          & 0.1711 (13) & 0.2368 (8)          & 0.0855 (12) \\
    UNED-GELP Run 1       & 0.4500 (5) & 0.8968 (8)          & 0.1625 (14) & 0.3168 (3)          & 0.1144 (6)  \\
    Awakened Run 2        & 0.5000 (3) & 0.8802 (12)         & 0.2625 (7)  & 0.3374 (2)          & 0.1709 (4)  \\
    \bottomrule
  \end{tabular}
  \normalsize
\end{table}

Latency-aware metrics reflect our trade-off between predictive depth and conversation length. DS@GT's LDCHR scores of 0.2091, 0.1696, and 0.2493 for Runs 1, 2, and 3 respectively placed the team mid-field and illustrate the cost of our moderately long interview strategy. Our system averaged 11.5 messages per conversation, well below the 30-message threshold at which the latency penalty reaches 0.5, but longer than faster teams like SINAI (4.4 messages) or Awakened (5.3 messages), which traded symptom coverage for speed. The LASHR scores of 0.1138, 0.1142, and 0.0777 similarly highlight the trade-off. Future iterations could benefit from a more aggressive early-stopping policy once high-confidence symptoms are locked in to preserve ADODL performance while recovering latency-weighted rank.

Overall, across ADODL rankings for complete submissions, DS@GT occupied ranks 11, 8, and 3 across its three runs, with the hybrid-generated Run 3 as the strongest individual submission. The fact that a run produced by the open-source Gemma 27B placed 3rd among all runs across all 21 complete-submission teams reinforces the efficacy of the hybrid design and algorithmic guidance. Comparable diagnostic accuracy is achievable at roughly one-quarter of the API cost with no manual intervention.

\section{Conclusion}

This paper presented an iterative, multi-agent LLM framework for conversational depression screening, developed and evaluated under the task constraints of no ground truth and weekly persona releases. Our system evolved through three stages. We began with a monolithic single-model prototype, progressed to a baseline multi-agent architecture, and arrived at an upgraded hybrid configuration.

The central hypothesis driving our final system was that a weaker, open-source model (Gemma 27B) could produce results competitive with a stronger paid model (GPT-5-nano) for the interviewer role, provided it received sufficient algorithmic supervision. Our results support this claim. Across 14 personas compared internally, the hybrid's mean BDI-II scores remained within 6 points of the baseline, approximately one severity category, for 13 of 14 personas. In the official eRisk evaluation, hybrid Run 3 achieved ADODL 0.9063, the 3rd highest score among all complete-submission runs and the strongest of our three submitted runs, outperforming the paid baseline Run 1 at 0.8841. This was accomplished at approximately \$2 per persona versus \$8 for the paid-only baseline, a 75\% cost reduction.

The primary limitation of our evaluation is the absence of a controlled ablation study. The baseline and hybrid configurations were introduced at different points in the persona sequence and continued to evolve through the submission period, meaning the two systems were never held constant against the same persona set under identical conditions. Compounding this, the unequal sample pools, 20 hybrid runs versus 10 baseline runs per persona, mean the two configurations are cost-matched but not sample-matched, and we cannot fully disentangle the contribution of sample count from that of the algorithmic components without an equal-pool replication. A rigorous follow-up experiment would freeze both systems, equalize sample counts, and evaluate across a fixed held-out persona set against released ground truth labels. Future studies could also benefit from tuning thresholds in the orchestration layer and cluster imputation to better balance symptom coverage against conversation length.

More broadly, however, these findings suggest that raw model capability is not a binding constraint in multi-agent conversational assessment. With structured orchestration, a question policy, and consensus-based aggregation, a 27B open-source model can compete with a near-frontier paid model. Hybrid systems have practical implications for cost- or privacy-sensitive deployments where proprietary API access is infeasible, such as clinical depression screening where patient data confidentiality is paramount. While our hybrid configuration still routes scoring through a proprietary model, the locally hosted interviewer model handles the full conversational exchange and therefore the most sensitive patient-facing interactions. A fully open-source pipeline, in which scoring is similarly migrated, remains a natural next step.

\begin{acknowledgments}
  We want to thank the Data Science at Georgia Tech (DS@GT) ARC group for their support and the eRisk 2026 organizing committee for designing and running this task. Computing resources were provided by the Georgia Institute of Technology in Atlanta, Georgia, USA, through access to the Partnership for an Advanced Computing Environment (PACE)~\cite{pace}. 

  Within the team, we acknowledge Murilo Gustineli for organizing team efforts, Anthony Miyaguchi for inspiring many ideas explored in this paper, and the rest of the DS@GT ARC team for their continued collaboration and support.
\end{acknowledgments}

\section*{Declaration on Generative AI}
During the preparation of this work, we used Claude (Anthropic) and ChatGPT (OpenAI) to check grammar and spelling, improve flow, and format LaTeX. After using these tools, we reviewed and edited the writing as needed and take full responsibility for the publication's content.

\bibliography{references}

\appendix

\section{Full Dialogue Tree}
\label{appendix:dialogue_tree}

The precomputed dialogue tree is organized into seven symptom clusters. The interview always opens with the \textit{Mood \& Pleasure} cluster (Cluster~1), where sadness and anhedonia are hardcoded as the initial focus symptoms. Subsequent openers are selected on demand. Each opener is indexed by its target BDI-II symptoms, and the system retrieves the first unused
opener whose targets appear in the orchestrator's current priority list of unassessed or low-confidence items. As the scorer gains confidence across turns, symptoms drop out of that list and their openers are skipped, so not all openers fire in every interview. Within each opener, BM25 keyword matching on the persona's response determines which follow-up branch
is taken. Follow-ups marked \textbf{[severity probe]} are designed to distinguish mild from moderate-to-severe presentations of the target BDI-II item. Rather than establishing mere presence of a symptom, they elicit information (e.g.\ frequency, duration, or functional impact) that maps onto BDI-II score levels 1-3.

\subsection*{Cluster 1: Mood \& Pleasure}
\textit{Target symptoms: sadness (Q1), anhedonia (Q4)}

\begin{description}[style=nextline, leftmargin=1.2em]
  \item[Sadness (Q1)]
    ``How have you been feeling in general lately? Has your mood been pretty steady, or has it
    been kind of all over the place?''
    \begin{description}[style=nextline, leftmargin=1.2em]
      \item[Follow-up (negative response)]
        ``When you say you've been feeling down, is that most of the time lately, or just
        certain moments?'' \textbf{[severity probe]}
      \item[Follow-up (positive response)]
        ``That's good to hear. Has anything been bringing you joy recently? Like hobbies or
        activities you look forward to?''
    \end{description}

  \item[Anhedonia (Q4)]
    ``What kinds of things do you find yourself enjoying these days? Hobbies, time with people,
    specific activities — what brings you a sense of satisfaction?''
    \begin{description}[style=nextline, leftmargin=1.2em]
      \item[Follow-up (negative response)]
        ``So it sounds like things that used to feel good don't really anymore. How long has
        that been going on?'' \textbf{[severity probe]}
      \item[Follow-up (positive response)]
        ``That sounds really nice. When you think about what's coming up — next week, next
        month — how do you generally feel about it?''
    \end{description}
\end{description}

\subsection*{Cluster 2: Physical}
\textit{Target symptoms: energy loss (Q15), sleep disturbance (Q16), appetite change (Q18), fatigue (Q20)}

\begin{description}[style=nextline, leftmargin=1.2em]
  \item[Energy Loss \& Fatigue (Q15, Q20)]
    ``How's your energy level been? Are you able to tackle tasks and activities, or do you find
    yourself dragging?''
    \begin{description}[style=nextline, leftmargin=1.2em]
      \item[Follow-up (negative response)]
        ``When you say you're exhausted, does that make it hard to do everyday things, or is
        it more that you don't feel the motivation to start?'' \textbf{[severity probe]}
      \item[Follow-up (positive response)]
        ``Good to know. How has your sleep been? Sleeping okay, having trouble, or sleeping
        more than usual?''
    \end{description}

  \item[Sleep Disturbance (Q16)]
    ``How has your sleep been? Are you sleeping okay, having trouble falling asleep, or waking
    up too early?''
    \begin{description}[style=nextline, leftmargin=1.2em]
      \item[Follow-up (negative response)]
        ``Has that always been an issue for you, or is this something that's changed
        recently?'' \textbf{[severity probe]}
    \end{description}

  \item[Appetite Change (Q18)]
    ``Have you noticed any changes in your appetite or eating habits lately? More, less, or
    about the same?''
    \begin{description}[style=nextline, leftmargin=1.2em]
      \item[Follow-up (negative response)]
        ``Is that change something that's been gradually getting worse, or did it happen
        pretty suddenly?'' \textbf{[severity probe]}
    \end{description}
\end{description}

\subsection*{Cluster 3: Outlook \& Social}
\textit{Target symptoms: pessimism (Q2), loss of interest (Q12)}

\begin{description}[style=nextline, leftmargin=1.2em]
  \item[Pessimism (Q2)]
    ``When you think about what's coming up — whether it's next week, next month, or further
    ahead — how do you generally feel about it?''
    \begin{description}[style=nextline, leftmargin=1.2em]
      \item[Follow-up (negative response)]
        ``It sounds like you're not feeling very optimistic. Has it always been like this, or
        is it newer?'' \textbf{[severity probe]}
    \end{description}

  \item[Loss of Interest (Q12)]
    ``Have you been spending time with people, or has that kind of taken a back seat? How's
    your social life feeling?''
    \begin{description}[style=nextline, leftmargin=1.2em]
      \item[Follow-up (negative response)]
        ``Is it that you're avoiding people by choice, or is it more like you just don't have
        the energy or interest?'' \textbf{[severity probe]}
    \end{description}
\end{description}

\subsection*{Cluster 4: Cognition}
\textit{Target symptoms: indecisiveness (Q13), concentration difficulty (Q19)}

\begin{description}[style=nextline, leftmargin=1.2em]
  \item[Concentration \& Indecisiveness (Q19, Q13)]
    ``Do you find it easy to focus on things — work, reading, conversations? Or has it been
    harder to concentrate lately?''
    \begin{description}[style=nextline, leftmargin=1.2em]
      \item[Follow-up (negative response)]
        ``So focusing feels harder. Is that affecting work or relationships or day-to-day
        stuff?'' \textbf{[severity probe]}
    \end{description}
\end{description}

\subsection*{Cluster 5: Self-Perception}
\textit{Target symptoms: self-dislike (Q7), feelings of worthlessness (Q14)}

\begin{description}[style=nextline, leftmargin=1.2em]
  \item[Self-Dislike \& Worthlessness (Q7, Q14)]
    ``How do you feel about yourself overall? Are you generally okay with who you are, or do
    you find yourself being quite critical?''
    \begin{description}[style=nextline, leftmargin=1.2em]
      \item[Follow-up (negative response)]
        ``That sounds like you're being pretty hard on yourself. Do you feel that way because
        of specific things, or is it more general?'' \textbf{[severity probe]}
    \end{description}
\end{description}

\subsection*{Cluster 6: Emotional Reactions}
\textit{Target symptoms: crying (Q10), agitation (Q11), irritability (Q17)}

\begin{description}[style=nextline, leftmargin=1.2em]
  \item[Crying (Q10)]
    ``Have you found yourself getting more emotional than usual? Like tearing up or crying more
    often?''
    \begin{description}[style=nextline, leftmargin=1.2em]
      \item[Follow-up (negative response)]
        ``Does that happen at certain times, or does it kind of come out of nowhere?''
        \textbf{[severity probe]}
      \item[Follow-up (positive response)]
        ``That's okay. Have you been feeling more on edge or restless than usual?''
    \end{description}

  \item[Irritability \& Agitation (Q17, Q11)]
    ``Have you noticed yourself getting annoyed or irritated more easily than usual? Like
    snapping at people or feeling on edge?''
    \begin{description}[style=nextline, leftmargin=1.2em]
      \item[Follow-up (negative response)]
        ``Is that something that's been building up, or did it come on pretty suddenly?''
        \textbf{[severity probe]}
    \end{description}
\end{description}

\subsection*{Cluster 7: Guilt \& Self-Criticalness}
\textit{Target symptoms: guilt (Q5), self-criticalness (Q8)}

\begin{description}[style=nextline, leftmargin=1.2em]
  \item[Guilt \& Self-Criticalness (Q5, Q8)]
    ``Do you find yourself feeling guilty a lot? About things you've done, or sometimes just in
    general?''
    \begin{description}[style=nextline, leftmargin=1.2em]
      \item[Follow-up (negative, pervasive guilt)]
        ``Is that guilt connected to something specific, or does it feel more pervasive?''
        \textbf{[severity probe]}
      \item[Follow-up (negative, self-criticism)]
        ``It sounds like you hold yourself to a pretty high standard. Do you feel like you're
        being too hard on yourself?'' \textbf{[severity probe]}
    \end{description}
\end{description}

\section{Cluster Imputation Donor Symptoms}
\label{appendix:imputation_donors}

When the scorer marks a symptom as unassessed (a confidence of zero), its score can be imputed from semantically related donor symptoms rather than defaulting to zero. Each imputable symptom is mapped to three ordered donors, weighted 0.5, 0.3, and 0.2 in priority order. A donor contributes to the estimate only if it was directly assessed with a confidence of at least 0.7 and a score of at least 2, and the resulting weighted mean is capped at a moderate severity of 2 so that imputation never assigns maximum severity. The four sensitive items (Q3 past failure, Q6 punishment feelings, Q9 suicidal thoughts, and Q21 loss of interest in sex) are never imputed and remain at zero unless directly assessed.

\begin{table}[htbp]
  \caption{Donor symptoms for cluster-based imputation. Each unprobed symptom is imputed from up to three directly assessed donors, listed in priority order with weights 0.5, 0.3, and 0.2. Donor entries are given as BDI-II item numbers. The four sensitive items (Q3, Q6, Q9, Q21) are never imputed.}
  \label{tab:imputation_donors}
  \begin{tabular}{lccc}
    \toprule
    Target symptom & Donor 1 (0.5) & Donor 2 (0.3) & Donor 3 (0.2) \\
    \midrule
    Q1 Sadness                   & Q4  & Q2  & Q10 \\
    Q2 Pessimism                 & Q1  & Q4  & Q14 \\
    Q4 Loss of pleasure          & Q1  & Q12 & Q20 \\
    Q5 Guilty feelings           & Q8  & Q7  & Q14 \\
    Q7 Self-dislike              & Q14 & Q8  & Q5  \\
    Q8 Self-criticalness         & Q5  & Q7  & Q14 \\
    Q10 Crying                   & Q1  & Q4  & Q5  \\
    Q11 Agitation                & Q17 & Q15 & Q1  \\
    Q12 Loss of interest         & Q4  & Q1  & Q20 \\
    Q13 Indecisiveness           & Q19 & Q20 & Q15 \\
    Q14 Worthlessness            & Q7  & Q8  & Q1  \\
    Q15 Loss of energy           & Q20 & Q16 & Q1  \\
    Q16 Changes in sleep         & Q20 & Q15 & Q18 \\
    Q17 Irritability             & Q11 & Q1  & Q20 \\
    Q18 Changes in appetite      & Q16 & Q20 & Q1  \\
    Q19 Concentration difficulty & Q13 & Q20 & Q15 \\
    Q20 Tiredness or fatigue     & Q15 & Q16 & Q18 \\
    \bottomrule
  \end{tabular}
\end{table}

\section{Interviewer Agent System Prompt}
\label{appendix:interviewer_prompt}

The following system prompt is injected at the start of every interviewer agent call. Each turn appends the orchestrator's current symptom focus and the full conversation history beneath it.

\begin{lstlisting}[basicstyle=\footnotesize\ttfamily, frame=single, breaklines=true]
# System Prompt: Empathetic Interviewer

You are a warm, empathetic conversational partner for the eRisk 2026 depression
detection task. Your ONLY job is to have a natural conversation with a persona.
You do NOT score or assess -- a separate system handles that.

## Critical Rules

1. Never mention depression, mental health, diagnoses, or clinical terms. This
   is a strict competition rule. Personas are designed to deflect such questions.
2. You speak first on the first turn with a warm, casual greeting.
3. Be natural and empathetic. You are a supportive conversationalist. Validate
   feelings. Reflect back what you hear. Transition smoothly between topics.
4. Output ONLY your message text. No JSON, no metadata, no reasoning. Just the
   words you would say to the persona.
5. Follow the guidance below. A separate system tells you which topics to
   explore. Work those topics into natural conversation -- never read them as a
   list.
6. One question at a time. You may combine 2-3 related topics into one natural
   question, but never rapid-fire a list of questions.
7. Keep messages concise. 2-4 sentences is ideal. Occasionally longer for
   validation + transition.

## Conversational Techniques

- Transitions: Link to what they just said. "You mentioned feeling tired -- I'm
  curious, how has your sleep been?"
- Validation: Brief acknowledgments: "That makes sense." / "That sounds really
  tough." / "Thank you for sharing that."
- Severity probing: When something is clearly present, ask ONE follow-up about
  frequency, duration, or impact. Pick one angle, don't stack questions.
- Combo questions: Naturally combine related topics: "How have the basics been
  -- sleep, energy, appetite?" This covers more ground per turn.
- Open-ended phrasing: "How has X been for you?" is better than "Do you have
  trouble with X?"

## Wrapping Up

When the guidance says to wrap up, thank the persona warmly and close the
conversation naturally. Do NOT ask any new questions during wrap-up.

## What You Receive Each Turn

Below, you will see:
1. Current Focus -- guidance on which topics to explore or whether to wrap up
2. Conversation So Far -- the full conversation history

Generate your next message based on these.
\end{lstlisting}

\section{Scorer Agent System Prompt}
\label{appendix:scorer_prompt}

The scorer agent receives this system prompt plus the full conversation transcript after each interviewer turn. It returns a structured JSON object with per-item scores, confidence values, and evidence quotes for all 21 BDI-II items.

\begin{lstlisting}[basicstyle=\footnotesize\ttfamily, frame=single, breaklines=true]
# System Prompt: BDI-II Transcript Scorer

You are a clinical assessment specialist. You will receive a conversation
transcript between an interviewer and a persona. Your job is to analyze the
transcript and score all 21 items of the Beck Depression Inventory-II (BDI-II).

## Instructions

1. Read the entire transcript carefully.
2. For each of the 21 BDI-II items, determine:
   - score (0-3): Based on the rubric for that item
   - confidence (0.0-1.0): How confident you are in this score
   - evidence: Quote or paraphrase the transcript text that supports your score
3. Think through each symptom carefully before assigning a score.
4. If the transcript provides NO evidence for a symptom, score it 0 with
   confidence 0.0 and evidence "No relevant discussion in transcript."
5. If the transcript provides SOME evidence but it's ambiguous, assign your
   best estimate with confidence 0.2-0.5 and explain the ambiguity.
6. Base scores ONLY on what the persona actually said. Do not infer beyond
   the text.
7. Frequency mapping:
   - "sometimes" / "occasionally" -> score 1
   - "a lot" / "most of the time" / "every hour or two" -> score 2
   - "all the time" / "constantly" / "all day" / "every day" -> score 3
8. Emotional expressions count: treat emotional descriptions (wanting to smash
   things, feeling rage, lump in throat) as direct evidence.

## BDI-II Scoring Rubric (abbreviated)

q01_sadness: 0=none; 1=occasional/mild; 2=frequent/pervasive (daily);
  3=overwhelming/unbearable. Daily or constant sadness = score 2 minimum.
q02_pessimism: 0=neutral/positive; 1=some discouragement; 2=clear pessimism;
  3=complete hopelessness.
q03_past_failure: 0=none; 1=some regrets; 2=pattern of failure thinking;
  3=pervasive failure identity.
q04_anhedonia: 0=enjoys as before; 1=reduced enjoyment; 2=very little pleasure;
  3=complete loss. If any enjoyment remains, cap at 2.
q05_guilt: 0=none; 1=some guilt; 2=frequent guilt; 3=constant/pervasive.
  "It's constant" or "I can't escape it" = score 3.
q06_punishment: 0=none; 1=vague sense of deserving; 2=expects punishment;
  3=feels actively punished.
q07_self_dislike: 0=normal; 1=lost confidence; 2=disappointed in self;
  3=active self-dislike/hatred.
q08_self_criticalness: 0=normal; 1=more self-critical; 2=criticizes self for
  all faults; 3=blames self for everything.
q09_suicidal_thoughts: 0=none; 1=thoughts but wouldn't act; 2=desire to die;
  3=would act if possible. Only score >0 if spontaneously mentioned.
q10_crying: 0=none; 1=occasionally; 2=frequently/easily triggered;
  3=constant/uncontrollable. Urge to cry counts equally as crying.
q11_agitation: 0=none; 1=somewhat on edge; 2=significant agitation/outbursts;
  3=constant/uncontrollable. Includes emotional agitation (rage, wanting to
  break things).
q12_loss_of_interest: 0=none; 1=somewhat less interested; 2=lost most interest;
  3=can't be interested in anything.
q13_indecisiveness: 0=normal; 1=somewhat harder; 2=much greater difficulty;
  3=can't make any decisions.
q14_worthlessness: 0=none; 1=feels less worthwhile; 2=feels worthless vs others;
  3=utterly worthless. Global statements of having no value = score 3.
q15_energy: 0=normal; 1=less than usual; 2=not enough for much; 3=no energy.
  "I've lost my energy for life" = score 3.
q16_sleep: 0=no change; 1=somewhat more/less; 2=much more/less (mind races);
  3=extreme disruption. "I can't relax" = score 2 minimum.
q17_irritability: 0=normal; 1=more irritable; 2=much more/snaps at people;
  3=irritable all the time.
q18_appetite: 0=no change; 1=somewhat changed; 2=much changed;
  3=extreme change. Complete loss or uncontrollable eating = score 3.
q19_concentration: 0=normal; 1=somewhat harder; 2=hard to maintain focus;
  3=can't concentrate on anything. "Brain goes blank" = score 2-3.
q20_fatigue: 0=normal; 1=tires more easily; 2=too tired for many activities;
  3=too tired for most. "Wake up tired, work tired, come home tired" = score 3.
q21_sex: 0=no change; 1=somewhat less; 2=much less; 3=complete loss.
  Only score >0 if spontaneously mentioned.

## Output Format

Output a JSON object with this exact structure. Include ALL 21 items. Output
ONLY the JSON -- no additional text before or after.

{
  "reasoning": "2-3 sentence overall assessment summary",
  "scores": {
    "q01_sadness": {
      "score": 2,
      "confidence": 0.85,
      "evidence": "Persona said 'I feel sad most of the time'..."
    },
    ...
  }
}
\end{lstlisting}

\section{Orchestrator Guidance Injection}
\label{appendix:orchestrator_guidance}

After each scorer response, the orchestrator deterministically constructs a natural-language guidance block that is prepended to the next interviewer call. The three methods below produce all prompt text the orchestrator injects: the opening turn, the mid-interview focus block, and the wrap-up signal.

\begin{lstlisting}[language=Python, basicstyle=\footnotesize\ttfamily, frame=single, breaklines=true]
def _initial_guidance(self, turn: int) -> OrchestratorGuidance:
    return OrchestratorGuidance(
        focus_symptoms=["q01_sadness", "q04_anhedonia"],
        guidance_text=(
            "## Current Focus\n\n"
            "This is the start of the interview. Open with a warm greeting "
            "and ask broadly about mood and what the persona has been enjoying "
            "lately. Target: Sadness (q01) and Loss of Pleasure (q04).\n"
        ),
        wrap_up=False,
        turn_number=turn,
    )

def _build_guidance_text(self, scores, focus_symptoms, turn, stable,
                          lf_signals=None) -> str:
    lines = ["## Current Focus\n"]
    if focus_symptoms:
        lines.append("The following symptoms need more exploration:")
        for sid in focus_symptoms:
            s = scores.symptoms[sid]
            name = SYMPTOM_ID_TO_NAME[sid]
            if not s.assessed:
                lines.append(f"- **{name}** ({sid}): NOT YET ASSESSED")
            else:
                lines.append(
                    f"- **{name}** ({sid}): confidence {s.confidence:.1f}, "
                    f"current score {s.score}"
                )
        lines.append("")
    well_assessed = [
        s for s in scores.symptoms.values()
        if s.confidence >= self.confidence_threshold and s.assessed
    ]
    if well_assessed:
        names = ", ".join(
            f"{s.name} ({s.confidence:.1f})"
            for s in sorted(well_assessed, key=lambda x: -x.confidence)[:6]
        )
        lines.append(f"Well-assessed (do NOT revisit): {names}\n")
    assessed_count = sum(1 for s in scores.symptoms.values() if s.assessed)
    min_assessed = int(len(scores.symptoms) * self.min_coverage)
    if assessed_count < min_assessed:
        lines.append(
            f"**Coverage: {assessed_count}/{len(scores.symptoms)} symptoms "
            f"assessed (need {min_assessed}). Do NOT wrap up yet.**\n"
        )
    remaining = self.max_turns - turn
    if remaining <= 4:
        lines.append(
            f"**Turn budget low ({remaining} turns remaining).** "
            "Focus on the most important gaps.\n"
        )
    if stable:
        lines.append(
            "Scores have been stable. Consider wrapping up if you feel "
            "you have enough information.\n"
        )
    if lf_signals:
        detected = [
            SYMPTOM_ID_TO_NAME.get(s["symptom_id"], s["symptom_id"])
            for s in lf_signals
        ]
        if detected:
            lines.append(
                f"Detected in last response: {', '.join(detected)}. "
                "Consider following up on these.\n"
            )
    return "\n".join(lines)

def _wrap_up_text(self, reason: str) -> str:
    reasons = {
        "max_turns_reached": "Maximum turns reached.",
        "all_symptoms_confident": "All assessable symptoms are well-assessed.",
        "scores_stable": "Scores have stabilized across recent turns.",
    }
    return (
        "## Wrap Up\n\n"
        f"It is time to end the interview. "
        f"Reason: {reasons.get(reason, reason)}\n"
        "Thank the persona warmly for their openness and end the conversation "
        "naturally. Do NOT ask any new questions.\n"
    )
\end{lstlisting}

\end{document}